%% file: arxiv2.tex
\documentclass{article}
\usepackage{ijcai21}

\usepackage{times}
\usepackage{soul}
\usepackage{url}
\usepackage[hidelinks]{hyperref}
\usepackage[utf8]{inputenc}
\usepackage[small]{caption}
\usepackage{graphicx}
\usepackage{amsmath}
\usepackage{amsthm}
\usepackage{booktabs}
\usepackage{algorithm}
\usepackage{algorithmic}
\usepackage{times}

\usepackage{soul}
\usepackage{url}
\usepackage[hidelinks]{hyperref}
\usepackage[utf8]{inputenc}
\usepackage{graphicx}
\usepackage{amsmath}
\usepackage{xspace}
\usepackage{caption}
\usepackage{bbm}
\usepackage{multicol}
\usepackage{tabularx}
\usepackage{hyperref}
\usepackage{cite}
\usepackage{amsmath,amssymb,amsfonts}
\usepackage{algorithmic}
\usepackage{graphicx}
\usepackage{textcomp}
\usepackage{subcaption}
\usepackage{appendix}
\usepackage[T1]{fontenc}
\usepackage{float}
\usepackage{hyperref}
\usepackage{xspace}
\usepackage{graphicx}
\usepackage{titlesec}
\usepackage{amsmath}
\usepackage{tablefootnote}
\usepackage{footnote}
\usepackage{multirow}
\usepackage{wrapfig}
\usepackage{boldline}
\usepackage{fixltx2e}
\usepackage[hang,flushmargin]{footmisc}
\allowdisplaybreaks
\usepackage{booktabs}
\makeatletter
    \let\@internalcite\cite
    \def\cite{\def\citeauthoryear##1##2{##1, ##2}\@internalcite}
    \def\shortcite{\def\citeauthoryear##1{##2}\@internalcite}
    \def\@biblabel#1{\def\citeauthoryear##1##2{##1, ##2}[#1]\hfill}
\makeatother
\usepackage{multibib}
\newcites{SM}{References}

\newcommand*\samethanks[1][\value{footnote}]{\footnotemark[#1]}

\newcommand{\modelname}{{\sc Tango}\xspace}

\title{TANGO: Commonsense Generalization in Predicting Tool Interactions for Mobile Manipulators}

\usepackage{fancyhdr}
\fancypagestyle{plain}{ %
  \fancyhf{} 
  
}
\author{
Shreshth Tuli\thanks{\footnotesize 
Indicates equal contribution. \\Project Page: \url{https://github.com/reail-iitd/tango}.
}
$^{1, 2}$\and
Rajas Bansal\samethanks[1]$^{1}$\and
Rohan Paul$^1$\And
Mausam$^1$\\
\affiliations
$^1$Department of Computer Science and Engineering. Indian Institute of Technology Delhi, India\\
$^2$Department of Computing, Imperial College London, UK\\
}

\begin{document}

\maketitle
\thispagestyle{fancy}
\begin{abstract}
Robots assisting us in factories or homes must learn to make use of objects as tools to perform tasks, 
e.g., a tray for carrying objects. 
We consider the problem of learning commonsense knowledge of when a tool may be useful 
and how its use may be composed with other tools to accomplish 
a high-level task instructed by a human. 
We introduce a novel neural model, termed \modelname, for predicting task-specific tool interactions, trained using demonstrations from human teachers instructing a virtual robot 
in a physics simulator.
\modelname encodes the world state, comprising objects and 
symbolic relationships between them, using a graph neural network. 
The model learns to attend over the scene using knowledge of the goal 
and the action history, finally decoding the symbolic action to execute. 
%
%
Crucially, we address generalization to unseen environments 
where some known tools are missing, but alternative unseen tools are present. 
We show that by augmenting the representation of the environment with pre-trained embeddings 
derived from a knowledge-base, the model can generalize effectively to novel environments.
%
Experimental results show a 
60.5-78.9\% absolute improvement over the baseline in predicting successful symbolic plans in unseen settings 
for a simulated mobile manipulator. 
%
%


\end{abstract}

\input{introduction-ver-1.3}

\input{related-works-ver-1.1}

\input{problem-formulation-ver-1.5}
\input{technical-approach-ver-1.3}
\input{evaluation-ver-1.4}

\input{conclusions}
\clearpage

\bibliographystyle{named}
\bibliography{example}

\appendix

\section*{Appendix}

\section{Model and Training Details} 
We detail the hyper-parameters for the {\sc Tango} architecture (\underline{T}ool Inter\underline{a}ction Prediction \underline{N}etwork for \underline{G}eneralized \underline{O}bject environments) introduced 
in this paper. 

\emph{Graph Structured World Representation. }
The Gated Graph Convolution Network (GGCN) was implemented with $4$-hidden layers, each of size $128$, with convolutions across $2$ time steps for every relation passing through a layer normalized GRU cell. 
The Parameterized ReLU activation function with a $0.25$ negative input slope was used in all hidden layers. 
%

\emph{Word Embeddings. }
The word embeddings (derived from $\mathrm{ConceptNet}$) were of size $300$. 
Additionally, the semantic state of each object was encoded as a one-hot vector of size $29$. Typically, there were 
$35$ and $45$ objects in the home and factory domains respectively. 

%
\emph{Fusing Metric Information. }
The metric encodings were generated from the metric information associated with objects using a $2$-layer Fully Connected Network (FCN) with $128$-sized layers. 

%
\emph{Encoding Action History. } 
A Long Short Term Memory (LSTM) layer of size $128$ was used to encode the action history using the generalized action encoding $\mathcal{A}(I_t(o^1_t, o^2_t))$. 

%
\emph{Goal-conditioned Attention. }
The attention network was realized as a $1$-layer FCN of layer size $128$ with a $\mathrm{softmax}$ layer at the end. 

\emph{Action Prediction. }
To predict the action $I_t$, a $3$-layer FCN was used, each hidden layer with size $128$ and output layer with size $|\mathcal{I}|$. 
$I_t$ was converted to a one-hot encoding $\vec{I}_t$. This, with the object embedding $e_o$ was passed to the $o^1_t$ predictor via an FCN. This FCN consists of 3-hidden layers of size $128$ and a final layer of size $1$ with a sigmoid activation (for likelihood). 
The $\vec{I}_t$ and $o^1_t$ likelihoods were sent to the $o^2_t$ predictor to predict likelihoods for all object embeddings 
$e_o$. 
This part was realized as a $3$-layer FCN with hidden layer size $128$ and final layer of size $1$ with a sigmoid activation function. 

\emph{Training parameters.} 
Model training used a learning rate of $5\times 10^{-4}$. 
The Adam optimizer~\cite{kingma2014adam} with a weight decay parameter of $10^{-5}$ and a batch size of $1$ was used. 
An early stopping criterion was applied for convergence. 
The \emph{action prediction accuracy} was used as the comparison metric on the validation set or up to a maximum of $200$ epochs.

\end{document}

%% file: introduction-ver-1.3.tex
\section{Introduction}\label{sec:introduction}
%
Advances in autonomy have enabled robots to enter human-centric domains
such as homes and factories where we envision them performing general purpose 
tasks such as transport, assembly, and clearing. 
Such tasks require a robot to interact with objects, 
often using them as \emph{tools}. 
For example, a robot asked to "take fruits to the kitchen", 
can use a \emph{tray} for carrying items, a \emph{stick} to fetch objects 
beyond physical reach and may use a \emph{ramp} to
reach elevated platforms.
We consider the problem of predicting \emph{which} objects 
could be used as tools and \emph{how} their use can be 
composed for a task.  
%

\begin{figure}[t]
    \centering
    \setlength{\belowcaptionskip}{-10pt}
    \includegraphics[width=\columnwidth]{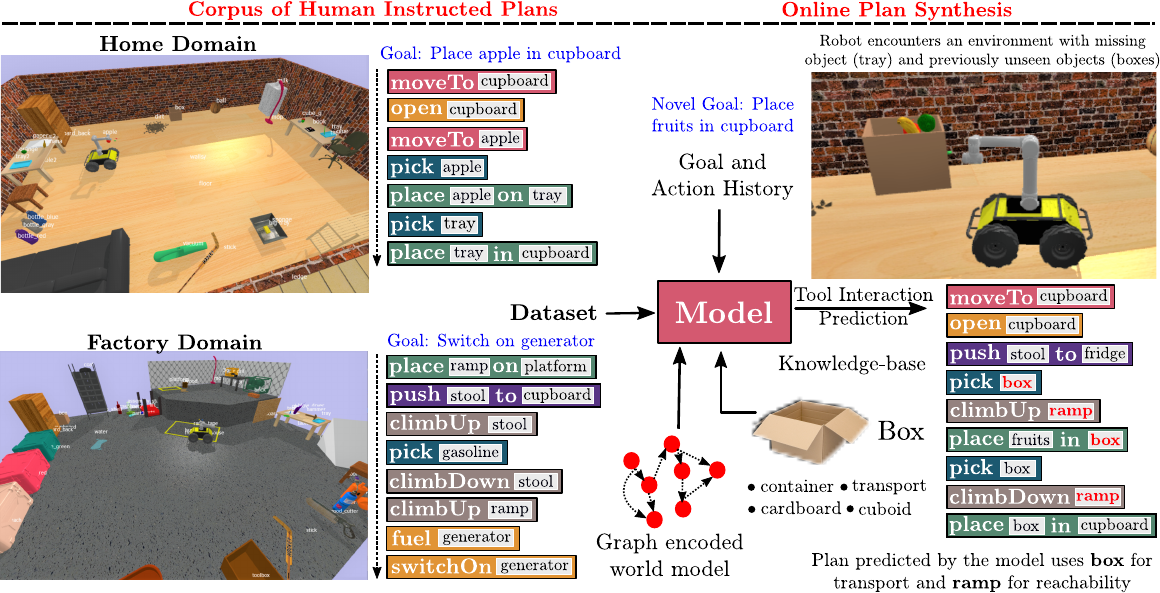}
    \caption{\footnotesize{
    \modelname acquires commonsense knowledge from human demonstrations leveraging graph-structured world representation, knowledge-base corpora and goal-conditioned attention to perform semantic tasks. Our aim is to acquire commonsense knowledge to develop a generalized goal-conditioned policy for a robot.
    %
    }
    }
    \label{fig:motivation}
\end{figure}

%
Learning to predict task-directed tool interactions poses several challenges. 
First, real environments (a household or factory-like domain) 
are typically large where an expansive number of tool interactions may be possible 
(e.g., objects supporting others while transporting).  
%
Acquiring data for all feasible tool objects or exploring the space of tool interactions 
is challenging for any learner. 
Second, the robot may encounter new environments populated with novel objects 
not encountered during training. 
Hence, the agent's model must be able to \emph{generalize} by reasoning 
about interactions with novel objects unseen during training.  
Humans possess innate commonsense knowledge about contextual use of tools for an intended goal~\cite{allen2019tools}. 
For example, a human actor when asked to move objects is likely to use trays, boxes, or even improvise with a new object with a flat surface. 
We aim at providing this commonsense to a robotic agent, so that it can generalize its knowledge to unseen tools, based on shared context and attributes of seen tools (see Figure~\ref{fig:motivation}). 

We acquire a data set of robot plans, where a human teacher guides a 
simulated mobile manipulator to perform tasks involving multi-step use of objects as tools. 
The model encodes the world state using a graph neural network 
and learns to attend over the scene using knowledge of the goal and the action history, 
finally decoding the symbolic action to execute. 
%
The model learns a dense representation of the object-centric graph of the environment which is 
augmented with word embeddings from a knowledge base, facilitating generalization to novel environments. 
The action predictions are interleaved with physics simulation (or execution) steps 
which ameliorates the need for modeling the complex effects of actions inherent in tool interactions.   
We term the model, \underline{T}ool Inter\underline{a}ction Prediction \underline{N}etwork for 
\underline{G}eneralized \underline{O}bject environments (\modelname).
Experimental evaluation with a simulated 
mobile manipulator demonstrate (a) accurate prediction of a tool interaction 
sequences with high executability/goal-attainment likelihood,
(b) common sense generalization to novel scenes with unseen object instances,  
and (c) robustness to unexpected errors during execution.  
%
%

%% file: related-works-ver-1.1.tex
\section{Related Work}\label{sec:related-works}
%
Learning control policies for manipulating tools has received recent attention in 
robotics. 
\cite{finn2017one} learn tool manipulation policies from  
human demonstrations.  
\cite{holladay2019force} learn physics-based models and 
\emph{effects} enabling compositional tool use.  
%
%
\cite{toussaint2018differentiable} present a planner to compose physics tool interactions 
using a logic-based symbolic planner. 
The aforementioned works focus on learning \emph{how} to manipulate 
a tool. In contrast, we extend our prior work on predicting 
\emph{which} objects may serve as tools for a given task \cite{toolnet} to generate plans. 

Others address the problem of acquiring knowledge for 
completing high-level task specifications. 
\cite{puig2018virtualhome,liao2019synthesizing} 
create a knowledge base of task decompositions as \emph{action sketches} and learn to translate 
sketches to executable plans. These efforts rely on the causal knowledge of sequences on 
sub-steps required to achieve an activity which are then contextually grounded. 
Instead, this work learns compositional tool use required to achieve the task without any
causal sequence as input.  
\cite{huang2019neural} learn task decompositions from human demonstration videos. 
However, the work does not explicitly model the physical constraints of the robot and does not 
generalize to new environments. 
\cite{boteanu2015towards} present a symbolic system where a robot imitates a demonstrations from a single teacher.  In new environments, it adapts the plan by performing object replacements using ConceptNet relation edges. In contrast, this paper proposes a neural model trained using a corpus of multiple and varied demonstrations provided by several teachers. Our model uses a dense embedding of semantic concepts, enabling generalization beyond relationships explicitly stored in ConceptNet. 

%
In the context of robot instruction following, 
\cite{nyga2018grounding} and \cite{kho2014robo} use curated knowledge bases to infer missing portions in 
instructions. Other such as \cite{jain2015planit} learn motion preferences 
implicit in commands. 
\cite{Bisk2020} learn \emph{physical} common sense knowledge 
for NLP tasks such as QA, analogy reasoning etc. 
The aforementioned approaches predict latent constraints or affordances for a 
specified task. This work, additionally predicts the \emph{sequence} of 
tool interactions implicitly learning the causal relationships between tools use and effects. 
\cite{misra2016tell} ground instructions for recipe preparation tasks. 
Their model can generalize to new recipes, but only in environments with previously \emph{seen} objects. 
In contrast, our model generalizes to worlds with previously \emph{unseen} tools.

%% file: problem-formulation-ver-1.5.tex
\section{Problem Formulation}\label{sec:formulation}
\textbf{Robot and Environment Model.} 
We consider a mobile manipulator operating in a known  
environment populated with objects. 
An object is associated with a pose, a geometric
model and symbolic states such as 
$\mathrm{Open/Closed}$, $\mathrm{On/Off}$ etc.  
We consider object interactions such as (i) \emph{support} e.g., 
a block supported on a tray, (ii) \emph{containment}: items placed 
inside a box/carton and (iii) \emph{attachment}: a nail attached to a wall, 
and (iv) \emph{contact}: a robot grasping an object.   
Let $s$ denote the world state that maintains (i) metric information: 
object poses, and (ii) symbolic information:  object states, class type and 
object relations as $\mathrm{OnTop}$, $\mathrm{Near}$, $\mathrm{Inside}$ and 
$\mathrm{ConnectedTo}$. 
Let $s_{0}$ denote the initial world state and 
$\mathcal{O}(\cdot)$ denote a map from world state $s$ to 
the set of object instances $O = \mathcal{O}(s)$ populating state $s$. 

Let $A$  denote the robot's symbolic action space. An action $a\in A$ is abstracted as  $I(o^1, o^2)$, with
an action type predicate $I \in \mathcal{I}$ that affects the states of objects 
$o^1 \in O$ and $o^2 \in O$, for instance, $\mathrm{Move(fruit_{0}, tray_{0})}$. 
We shall also use the notion of a timestamp as a subscript to indicate prediction for each
state in the execution sequence.
The space of robot interactions include grasping, releasing, pushing, 
moving an object to another location or inducing discrete state changes 
(e.g.. opening/closing an object, operating a switch or using a mop). 
%
%
We assume the presence of an underlying 
low-level metric planner, encapsulated as a robot \emph{skill}, which  
realizes each symbolic action or returns if the action is infeasible. 
Robot actions are stochastic, modeling 
execution errors (unexpected collisions) and unanticipated outcomes 
(objects falling, changing the symbolic state).
Let $\mathcal{T}(\cdot)$ denote the transition function.  
The successor state $s_{t+1}$ upon 
taking the action $a_{t}$ in state $s_{t}$ is sampled 
from a physics simulator. 
Let $\eta_{t} = \{ a_{0}, a_{1}, \dots, a_{t-1} \}$ denote the 
\emph{action history} till time $t$. 

The robot is instructed by providing a \emph{declarative} goal 
$g$ expressing the symbolic constraint between world objects.  
For example, the \emph{declarative} goal, "place milk in fridge" 
can be expressed as a constraint $\mathrm{Inside(milk_{0}, fridge_{0})}$ 
between specific object instances. 
Finally, the robot must synthesize a plan to satisfy the goal constraints.               
Goal-reaching plans may require using some objects 
as tools, for instance, using a container for moving items, 
or a ramp negotiate an elevation. 
%

%
\textbf{Predicting Tool Interactions.}  
Our goal is to aim at learning common sense knowledge 
about \emph{when} an object can be used as a tool 
and \emph{how} their use can be sequenced for goal-reaching plans. 
%
We aim at learning a policy $\pi$ 
that estimates the next action $a_{t}$ conditioned on the 
the goal $g$ and the initial state $s$ (including the action history $\eta_{t}$, 
such that the robot's \emph{goal-reaching} likelihood is maximized.  
We adopt the \textrm{MAXPROB-MDP} \cite{kolobov2012planning} formalism and 
estimate a policy that maximizes the goal-reaching likelihood from the given state
\footnote{\textrm{MAXPROB-MDP} can be equivalently viewed as 
an infinite horizon, un-discounted MDP with a zero reward for non-goal states and a 
positive reward for goal states \cite{kolobov-icaps11}.
}. 
Formally, let $P^{\pi}\left( s, g \right) $ denote the \emph{goal-probability} function  
that represents the likelihood of reaching the goal $g$ from a state $s$ on 
following $\pi$.
Let $S^{\pi_{s}}_{t}$ be a random variable denoting the state resulting from 
executing the policy $\pi$ from state $s$ for $t$ time steps. 
Let $\mathcal{G}(s,g)$ denote the Boolean \emph{goal check} function that determines  
if the intended goal $g$ is entailed by a world state $s$ as 
$\mathcal{G}(s,g) \in \{\mathrm{True(T)},\mathrm{False(F)} \}$. 
%
The policy learning objective is formulated as 
maximizing the likelihood of reaching a goal-satisfying state $g$ 
from an initial state $s_{0}$, denoted as $\max_{\pi} P^{\pi}(s_{0}, g)=$ 
\begin{gather*}
    \max_{\pi} 
    \sum_{t=1}^{\infty}  P \bigg( \mathcal{G} ( S^{\pi_{s_{0}}}_{t}, g )= \mathrm{T} :  \mathcal{G}( S^{\pi_{s_{0}}}_{t'}, g) = \mathrm{F},   
    ~\forall t' \in [1,t)\bigg). 
\end{gather*}
%
 
%
The policy is modeled as a function $f_{\theta}(.)$ parameterized by parameters 
$\theta$ that determines the next action for a given world state, the robot's action history and the 
goal as $a_{t} = f_{\theta} \left( s_{t}, g, \eta_{t} \right) $. 
We adopt imitation learning approach and learn the function $f_{\theta}(.)$ from
demonstrations by human teachers. 
Let $\mathcal{D}_{\mathrm{Train}}$ denote the corpus of $N$ goal-reaching plans, 
\begin{equation*}
    \mathcal{D_{\mathrm{Train}}} = \{ (s_0^i,g^i,\{s_j^i,a_j^i\}) \mid i \in \{1,N\}, j \in \{0,t_{i}-1\} \}, 
    \label{eq:data set}
\end{equation*}
where the $i^{th}$ datum 
consists of the initial state $s^{i}_0$, the goal $g^{i}$ and a state-action sequence 
$\{ (s^{i}_{0}, a^{i}_{0}), \dots, (s^{i}_{t-1}, a^{i}_{t-1}) \}$ of length $t_{i}$. 
The set of human demonstrations elucidate common sense knowledge 
about \emph{when} and \emph{how} tools can be used for attaining provided goals.  
The data set $\mathcal{D}_{\mathrm{Train}}$ supervises an imitation loss 
between the human demonstrations and the model predictions,
resulting in learned parameters $\theta^{*}$.  
%
Online, the robot uses the learned model to sequentially predict actions and execute in the 
simulation environment till the goal state is attained. 
%
We also consider the \emph{open-world} case where the robot may 
encounter instances of \emph{novel} object categories \emph{unseen} in training, 
necessitating a \emph{zero-shot} generalization.  

%% file: technical-approach-ver-1.3.tex
\section{Technical Approach}\label{sec:model}
\modelname{} learns to predict the next robot action $a_{t}$,  
given the world state $s_{t}$, the goal $g$ and the 
action history $\eta_{t}$. \modelname is realized as a neural network model $f_{\theta}$ as follows:
%
%
%
\begin{equation*}
  a_{t} = f_{\theta} \left( s_{t}, g, \eta_{t} \right) 
 = 
 f^{act}_{\theta}  \left( f^{goal}_{\theta} \left( f^{state}_{\theta} \left( s_{t} \right), g,  f^{hist}_{\theta} \left( \eta_{t} \right) \right) \right)
\label{eqn:nn}
\end{equation*}
It adopts an object-centric graph representation, learning 
a state encoding that fuses metric-semantic information about 
objects in the environment via function $f^{state}_{\theta}\left(\cdot \right)$. 
The function $f^{hist}_{\theta} \left(\cdot \right)$ encodes the action history. 
The model learns to attend over the world state conditioned 
on the declarative goal and the history of past actions through $f^{goal}_{\theta}\left( \cdot \right)$.  
Finally, the learned encodings are decoded as the next action for the 
robot to execute via $f^{act}_{\theta}\left( \cdot \right)$.   
Crucially, the predicted action is grounded over an 
\emph{a-priori} unknown state and type of objects in the environment. 
The predicted action is executed in the environment and the  updated state action history is used for estimation 
at the next time step. 
The constituent model components are detailed next. 
%
\begin{figure*}[t]
    \centering
    \setlength{\belowcaptionskip}{-10pt}
    \includegraphics[width=0.82\textwidth]{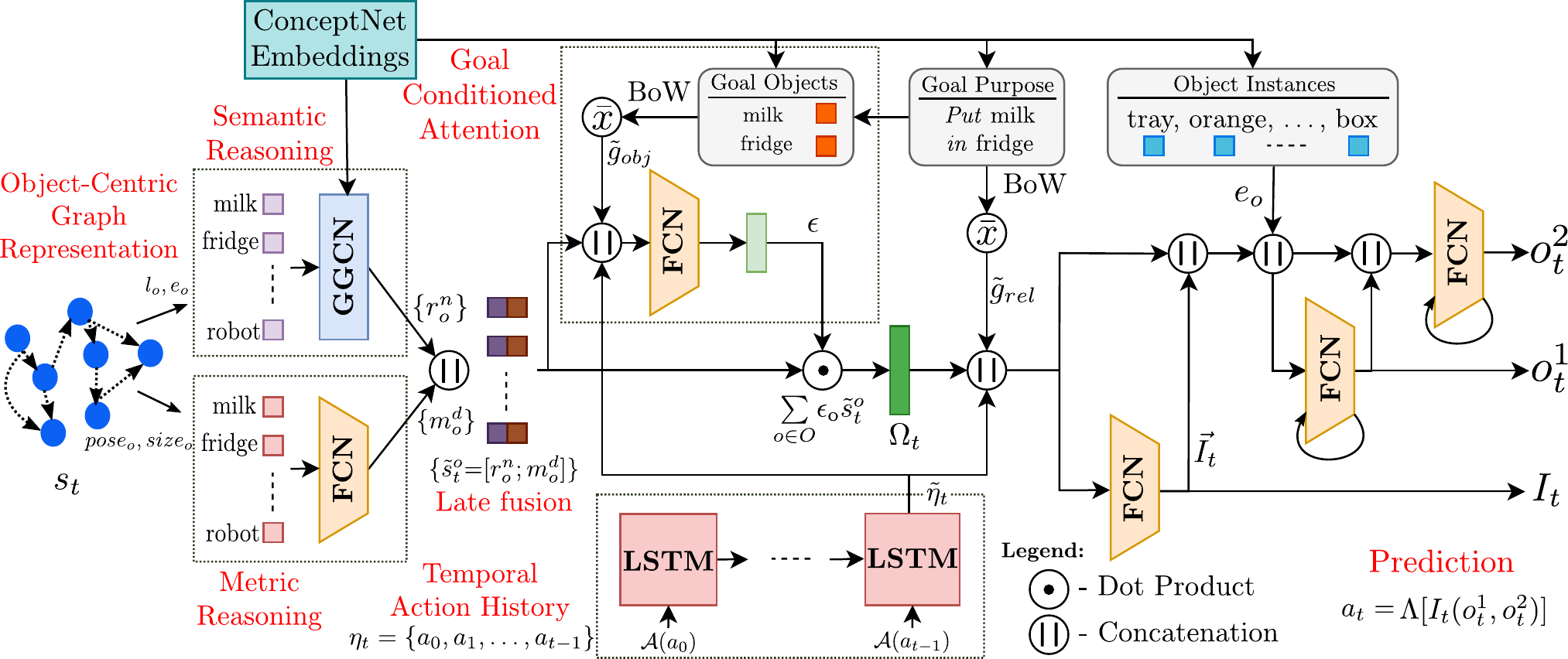}
  \caption{\footnotesize{ 
  \modelname neural model encodes the metric-semantic world state using graph convolution (GGCN) and fully connected (FCN) layers. The model uses goal information and the robot's action history to attend over a task-specific context, finally decoding the next symbolic action for the robot to execute. A graph structured representation and inclusion of pre-trained word embeddings (from a knowledge base) facilitate generalization in predicting interactions in novel contexts with new objects unseen in training. 
  }}    
    \label{fig:complete_model}
\end{figure*}

\vspace{0.5ex}
\noindent 
\textbf{Graph Structured World Representation. }
%
\modelname encodes a robot's current world state $s_t$  as an object-centric graph 
$G_t = (O, R)$. 
Each node in the graph represents an object instance $o \in O = \mathcal{O}(s_t)$.
The edge set consists of binary relationships $\mathrm{OnTop}$, $\mathrm{ConnectedTo}$, $\mathrm{Near}$ and $\mathrm{Inside}$ between objects $R \subseteq O \times O$.  
%
%
Let $l_{o} \in \{0,1\}^{p}$ represents discrete object states for the object $o$ (e.g. $\mathrm{Open/Closed}$, $\mathrm{On/Off}$).  
Next, it incorporates a pre-trained function $\mathcal{C}(\cdot)$ that embeds a 
word (like token of an object class or a relation) to a dense distributed representation, such that semantically close 
tokens appear close in the learned space \cite{mikolov2018advances}. 
The use of such embeddings enables generalization, which we discuss subsequently.  

%
Let $e_{o} = \mathcal{C}(o) \in \mathcal{R}^{q}$ denote the $q$-dimensional embedding for an object instance $o$. 
The embeddings $l_{o}$ and $e_{o}$ model object attributes that initialize the 
state of each object node in the graph. 
%
%
The local context for each $o$ is incorporated via 
a Gated Graph Convolution Network (GGCN)  \cite{liao2019synthesizing}, which
performs message passing between 1-hop vertices on the graph. 
Following \cite{puig2018virtualhome}, the gating stage is realized 
as a Gated Recurrent Unit (GRU) resulting in \emph{graph-to-graph} 
updates as: 
\begin{align*}
\begin{split}
    r_o^{0} &= \mathrm{tanh} \left (W_{r} \left[ l_{o}; e_{o} \right] + b^r \right),\\[-3pt]
    x^k_o &= \sum_{j \in R}\sum_{o' \in N^j(o)} W_j^k r_{o'}^{k-1} ,\\[-3pt]
    r^k_{o} &= \mathrm{GRU} \left( r^{k-1}_o, x^{k}_{o} \right).\\[-3pt]
\end{split}
\end{align*}
Here, the messages for object $o$ are aggregated over neighbors $N^j(o)$ connected by relation $j$ ($\forall j \in R$) 
during $n$ convolutions, resulting in an embedding $r^{n}_{o}$ for each object instance in the environment. 
%
%

\vspace{0.5ex}
\noindent 
\textbf{Fusing Metric Information. }
Next, \modelname incorporates the metric information associated with objects. 
Let $pose_o$ and $size_o$ represent the pose and size/extent (along xyz axes) 
for each object instance. 
The properties are encoded using a $d$-layer Fully Connected Network (FCN) with a 
Parameterized ReLU (PReLU) \cite{prelu} activation as: 
\begin{align*}
\begin{split}
	m^0_o &= \mathrm{PReLU} \left( W_{mtr}^{0}[pose_o; size_o] + b_{mtr}^0\right) \\[-3pt]
    m^k_o &= \mathrm{PReLU} \left( W_{mtr}^{k} m^{k-1}_o  +b_{mtr}^{k}\right), 
\end{split} 
\end{align*}
resulting in the metric encoding $m^d_o$ for each object in the scene. 
A world state encoding (for~$s_t$) is obtained by fusing 
the semantic and metric embeddings as $f^{state}_{\theta}(s_t) = \{\tilde{s}_t^o\! =\! [r^n_o;m^d_o]|\ \forall o \in {\cal O}(s_t)\}$.   
\emph{Late fusion} of the two encodings 
allows downstream predictors to exploit them independently.
 
\vspace{0.5ex}
\noindent 
\textbf{Encoding the Action History. }
The task of deciding the next action is informed by the agent's action history in two ways.
First, sequential actions are often temporally correlated. 
E.g., a placing task often involves moving close to the box, 
opening it and then placing an item inside it. 
Hence, maintaining the action history can help in prediction of the next action. 
Second, the set of actions the robot executed in the past 
provides a local context indicating the objects the 
robot may affect in future. 
Formally, \modelname encodes the temporal action history $\eta_t$
using an $\mathrm{LSTM}$ resulting in embedding vector $f^{hist}_{\theta}(\eta_t) = \tilde{\eta}_t$. 
We define action encoding $\mathcal{A}(a_{t-1})$ of 
$a_{t-1} = I_t (o^1_{t-1}, o^2_{t-1})$ independent of the object set, as 
$\mathcal{A}(a_{t-1}) = [\vec{I}_{t-1}; \mathcal{C}(o^1_{t-1}); \mathcal{C}(o^2_{t-1})]$, 
where $\vec{I}_{t-1}$ is a one-hot vector over possible interaction types 
$\mathcal{I}$, and 
$\mathcal{C}(o^1_{t-1})$ and $\mathcal{C}(o^2_{t-1})$ 
represent the word embeddings of the 
object instances $o^1_{t-1}$ and $o^2_{t-1}$. 
At each time step $t$, the $\mathrm{LSTM}$ 
encoder takes in the encoding of previous action, $\mathcal{A}(a_{t-1})$ and 
outputs the updated encoding $\tilde{\eta}_t$, given as  
$\tilde{\eta}_t = \mathrm{LSTM}(\mathcal{A}(a_{t-1}),\tilde{\eta}_{t-1})$.
%

\vspace{0.5ex}
\noindent 
\textbf{Goal-conditioned Attention. }
The goal $g$ consists of symbolic relations (e.g. $\mathrm{Inside}$, $\mathrm{OnTop}$ etc.) 
between object instances (e.g., carton and cupboard) that must be true at the end of the robot's plan execution. 
The declarative goal input to the model is partitioned as relations 
$g_{rel}$ and the object 
instances specified in the goal $g_{obj}$.
The resulting encondings are denoted as $\tilde{g}_{rel}$ and $\tilde{g}_{obj}$:
\begin{equation*}
    \tilde{g}_{rel} = \frac{1}{|g_{rel}|}  \sum_{j \in g_{rel}} \mathcal{C}(j) 
   \hspace{2mm} \mathrm{ and }    \hspace{2mm}
    \tilde{g}_{obj} = \frac{1}{|g_{obj}|} \sum_{o \in g_{obj}} \mathcal{C}(o).
\end{equation*}
Next, the goal encoding and the 
action history encoding $\tilde{\eta}_{t}$ is used to learn attention weights over objects in the 
environment $\epsilon_o = f^{goal}_\theta(\tilde{s}_{t}^o, \tilde{g}_{obj}, \tilde{\eta}_{t})$ \cite{bahdanau2014neural}.
This results in the attended scene encoding $\Omega_{t}$ as:
\begin{equation*}
\label{eqn:attn_scene}
    \Omega_{t} = \sum_{o \in O} \mathrm{\epsilon_o} \tilde{s}_t^o \mathrm{\ where,\ } 
    \epsilon_o = \mathrm{softmax} \left( W_g [ \tilde{s}_t^o; \tilde{g}_{obj}; \tilde{\eta}_t ]  + b_g \right).
\end{equation*}
The attention mechanism aligns the goal information with the scene learning a task-relevant context,  
relieving the model from reasoning about objects in the environment unrelated to the task, 
which may be numerous in realistic environments. 
\noindent 
\textbf{Robot Action Prediction. } 
\modelname takes the encoded information about the world state, goal and action history to decode the next symbolic action $a_t = I_t (o^1_t, o^2_t)$.
%
The three components $I_t$, $o^1_t$ and $o^2_t$ are predicted auto-regressively, where the prediction of the interaction, $I_t$ is used for the prediction of the first object, $o^1_t$ and both their predictions are used for the second object prediction, $o^2_t$.
The prediction is made using the encoding of the state, i.e the attended scene embedding $\Omega_t$, the relational description of the goal $\tilde{g}_{rel}$ and the action history encoding $\tilde{\eta}_t$.
For the object predictors $o^1_t$ and $o^2_t$, instead of predicting over a predefined set of objects, \modelname predicts a likelihood score of each object $o\in O$ based on its object embedding $ \tilde{s}_t^o$, and selects the object with highest likelihood score. 
The resulting factored likelihood allows the model to generalize to an \emph{a-priori} unknown number and types of object instances:   
%
\begin{gather*}
    I_t = \mathrm{argmax}_{I \in \mathcal{I}} \left( \mathrm{softmax}(W_I [\Omega_t; \tilde{g}_{rel}; \tilde{\eta}_t] + b_I) \right),\\
    \begin{split}
    o^1_t &= \mathrm{argmax}_{o \in {O}} \alpha_t^o \\ &= \mathrm{argmax}_{o \in {O}}\ ( \sigma(W_{\alpha} [\Omega_t; \tilde{g}_{rel}; \tilde{\eta}_t; e_o; \vec{I}_t] + b_{\alpha})) \mathrm{,}
    \end{split}\\
    o^2_t = \mathrm{argmax}_{o \in {O}}\ ( \sigma(W_{\beta} [\Omega_t; \tilde{g}_{rel}; \tilde{\eta}_t; e_o; \vec{I}_t; \alpha_t^o] + b_{\beta})).
\end{gather*}
Here $\alpha^o_t$ denotes the likelihood prediction of the first object. The model is trained using a Binary Cross-Entropy loss, with the loss for the three predictors being added independently. 
%
Finally, we impose grammar constraints (denoted as $\Lambda$) at inference time 
based on the number of arguments that the predicted interaction $I_t$ accepts. 
If $I_t$ accepts only one argument only $o^1_t$ is selected, otherwise 
both are used. 
%
%
Thus, predicted action, $a_t = f^{act}_{\theta}(\Omega_t,\tilde{g}_{rel},\tilde{\eta}_t) = \Lambda [ I_t (o^1_t, o^2_t) ]$, 
is then executed by the robot in simulation.
The executed action and resulting world state is provided as input to the model 
for predicting the action at the next time step. 
%
 
\vspace{0.5ex}
\noindent \textbf{Word Embeddings Informed by a Knowledge Base}
\modelname uses word embedding function $\mathcal{C}(\cdot)$ that provides a  
dense vector representation for word tokens associated with object class and relation types.  
%
%
Contemporary models
use word embeddings acquired from language modeling tasks \cite{mikolov2018advances}. 
We adopt embeddings that are additionally informed by an existing knowledge graph 
$\mathrm{ConceptNet}$ \cite{speer2017conceptnet} that 
 provides a sufficiently large knowledge graph connecting words with edges expressing relationships such as 
$\mathrm{SimilarTo}$, $\mathrm{IsA}$, $\mathrm{UsedFor}$, $\mathrm{PartOf}$ and $\mathrm{CapableOf}$. 
Word embeddings \cite{mikolov2018advances} can be \emph{retro-fitted} 
such that words related using knowledge graph embeddings are also close in the embedding space \cite{conceptnet-github}. 
Using such (pre-trained) embeddings incorporates \emph{general purpose} relational 
knowledge to facilitate richer generalization for downstream policy learning. 
The complete sequence of steps is summarized in Figure~\ref{fig:complete_model}. 

%% file: evaluation-ver-1.4.tex
\section{Data Collection Platform and Annotation}
\begin{figure}[]
    \centering
    \resizebox{0.8\columnwidth}{!}{
    \begin{tabular}{|c|}
    \hline 
    \textbf{Robot Actions}\tabularnewline
    \hline 
    \begin{minipage}[t]{\columnwidth}Push, Climb up/down, Open/Close, Switch on/off, Drop, Pick, Move to, Operate device,
    Clean, Release material on surface, Push until force\end{minipage}\tabularnewline
    \hline 
    \hline 
    \textbf{Object Attributes}\tabularnewline
    \hline 
    \begin{minipage}[t]{\columnwidth}Grabbed/Free, Outside/Inside, On/Off, Open/Close, Sticky/Not Sticky, Dirty/Clean, Welded/Not Welded, Drilled/Not Drilled, Driven/Not Driven, Cut/Not Cut, Painted/Not Painted\end{minipage}\tabularnewline
    \hline 
    \hline
    \textbf{Semantic Relations}\tabularnewline
    \hline 
    On top, Inside, Connected to, Near\tabularnewline
    \hline 
    \hline
    \textbf{Metric Properties}\tabularnewline
    \hline 
    Position, Orientation, Size\tabularnewline
    \hline 
    \end{tabular}} \setlength{\belowcaptionskip}{-8pt}
    \captionof{table}{\footnotesize{Domain Representation. Robot symbolic actions, semantic attributes, relations to describe the world state and objects populating the scene in Home and Factory Domains. }}
    \label{tab:env_desc}
\end{figure}

\begin{table*}[]
    \centering
    \resizebox{\textwidth}{!}{
    \begin{tabular}{|c|c|c|c|c|c|}
    \hline 
    \multirow{2}{*}{Domain} & \multirow{2}{*}{Plan lengths} & Objects interacted & Tools used & \multirow{2}{*}{Sample objects} & \multirow{2}{*}{Sample goal specifications}\tabularnewline
     &  &  with in a plan & in a plan &  & \tabularnewline
    \hline 
    \hline 
    Home & 23.25$\pm$12.65 & 4.12$\pm$1.97 & 0.93$\pm$0.70 & \begin{minipage}[t]{0.7\textwidth}floor$^{1}$, wall, fridge$^{123}$, cupboard$^{123}$, tables$^{1}$, couch$^{1}$, \textbf{big-tray}$^{1}$, \textbf{tray}$^{1}$, \textbf{book}$^{1}$,
        paper, cubes, light switch$^{4}$, bottle, \textbf{box}$^{2}$, fruits, \textbf{chair}$^{15}$, \textbf{stick}, dumpster$^{2}$, milk carton, shelf$^{1}$, \textbf{glue}$^{6}$, \textbf{tape}$^{6}$, \textbf{stool}$^{15}$, \textbf{mop}$^{8}$, \textbf{sponge}$^{8}$, \textbf{vacuum}$^{8}$, dirt$^{7}$, door$^{2}$\end{minipage} & \begin{minipage}[t]{0.4\textwidth} 1. Place milk in fridge, 2. Place fruits in cupboard, 3. Remove dirt from floor, 4. Stick paper to wall, 5. Put cubes in box, 6. Place bottles in dumpster, 7. Place a weight on paper, 8. Illuminate the room. \vspace{2pt}
        \end{minipage}\tabularnewline
    \hline 
    Factory & 38.77$\pm$23.17 & 4.38$\pm$1.85 & 1.44$\pm$0.97 & \begin{minipage}[t]{0.7\textwidth}
        floor$^{1}$, wall, \textbf{ramp}, worktable$^{1}$, \textbf{tray}$^{1}$, \textbf{box}$^{2}$, crates$^{1}$, \textbf{stick}, long-shelf$^{1}$, \textbf{lift}$^{1}$, cupboard$^{123}$, \textbf{drill}$^{4}$, \textbf{hammer}$^{49}$, \textbf{ladder}$^{5}$, \textbf{trolley}$^{2}$, \textbf{brick}, \textbf{blow dryer}$^{48}$,
        \textbf{spraypaint}$^{4}$, \textbf{welder}$^{4}$, generator$^{4}$, \textbf{gasoline}, \textbf{coal}, \textbf{toolbox}$^{2}$, \textbf{wood cutter}$^{4}$, \textbf{3D printer}$^{4}$, screw$^{9}$, nail$^{9}$, \textbf{screwdriver}$^{49}$, wood, platform$^{1}$, oil$^{7}$, water$^{7}$,
        board, \textbf{mop}$^{8}$, paper, \textbf{glue}$^{6}$, \textbf{tape}$^{6}$, assembly station, spare parts, \textbf{stool}$^{15}$\vspace{2pt}\end{minipage} & \begin{minipage}[t]{0.4\textwidth} 1. Stack crated on platform, 2. Stick paper to wall, 3. Fix board on wall, 4. Turn on the generator, 5. Assemble and paint parts, 6. Move tools to workbench, 7. Clean spilled water, 8. Clean spilled oil. \vspace{2pt}
        \end{minipage}\tabularnewline
    \hline 
    \end{tabular}} \setlength{\belowcaptionskip}{-6pt}
    \caption{\footnotesize{Dataset characteristics. The average plan lengths, number of objects interacted in plan and number of tools used in plans with object and goal sets for Home and Factory domains. Object positions were sampled using Gaussian distribution. Objects in bold can be used as tools. Legend:- $^{1}$:~surface, $^{2}$:~can open/close, $^{3}$:~container, $^{4}$:~can operate, $^{5}$:~can climb, $^{6}$:~can apply, $^{7}$:~can be cleaned, $^{8}$:~cleaning agent, $^{9}$:~can 3D print. Objects in bold can be used as tools. Stool/ladder are objects used to represent a tool for raising the height of the robot.}}
    \label{tab:dataset_complete}
\end{table*}

\begin{figure}
    \centering \setlength{\belowcaptionskip}{-6pt}
    \includegraphics[width=\linewidth]{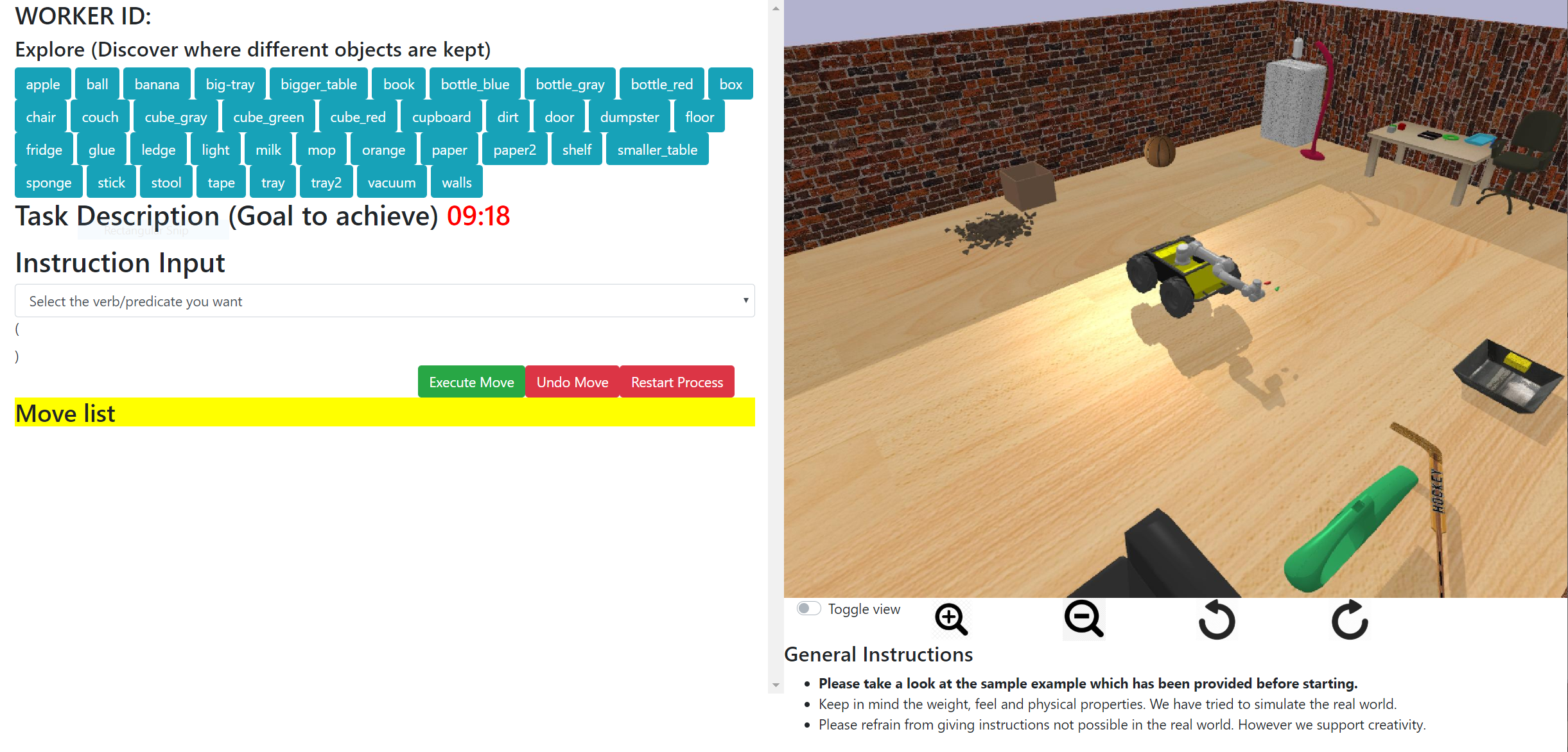}
    \caption{Data Collection Interface. The human teacher instructs (left) a virtual mobile manipulator robot by specifying symbolic actions. The human-instructed plan is simulated and visualized (right). }
    \label{fig:screenshot} 
\end{figure}

\textbf{Data Collection Environment.} 
%
We use PyBullet, a physics simulator~\cite{coumans2016pybullet}, 
to generate home and  factory-like environments  
populated with a virtual mobile manipulator (a Universal Robotics (UR5) arm 
mounted on a Clearpath Husky mobile base).
The robot is tasked with goals that involve multiple interactions with objects  
derived from standardized data sets \cite{calli2017ycb}. 
These goals include: 
(a) \emph{transporting} objects from one region to another (including space on top of or inside
other objects), 
(b) \emph{fetching} objects, which the robot must reach, grasp and return with, and 
(c) \emph{inducing state changes} such as illuminating the room or removing dirt from floor. 
Figure~\ref{fig:screenshot} illustrates the data collection platform. 
The set of possible actions and the range of interactions  are listed in Table~\ref{tab:env_desc}. 
The effects of actions such as pushing or moving are simulated via a motion planner and propagated to the next time step. 
Abstract actions such as attachment, operating a tool or grasping/releasing objects are encoded symbolically 
as the establishment or release constraints. {The simulation for these actions is coarse and considers their symbolic effects 
forgoing the exact motion/skills needed to implement them. We assume that the robot can realize abstract actions through low-level routines. }

\begin{figure}[]
    \centering
    \begin{subfigure}{0.48\linewidth}
        \centering
        \includegraphics[width=0.9\linewidth]{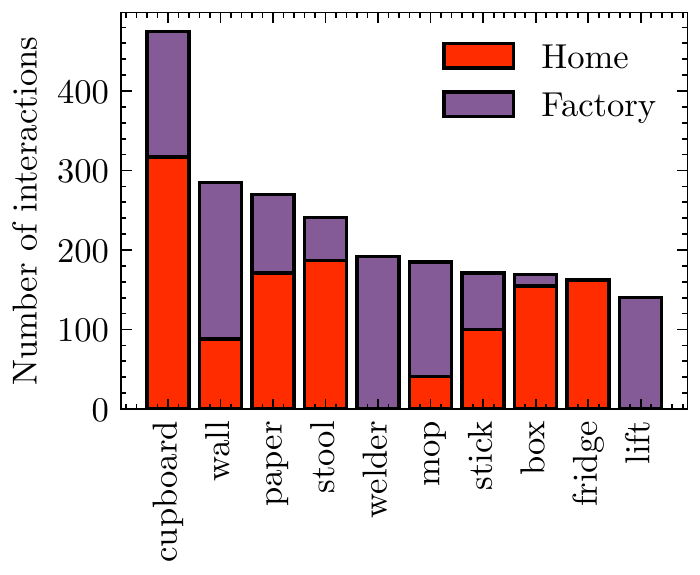}
        \caption{Object interactions}
        \label{fig:num_int}
    \end{subfigure}
    \begin{subfigure}{0.48\linewidth}
        \centering
        \includegraphics[width=0.9\linewidth]{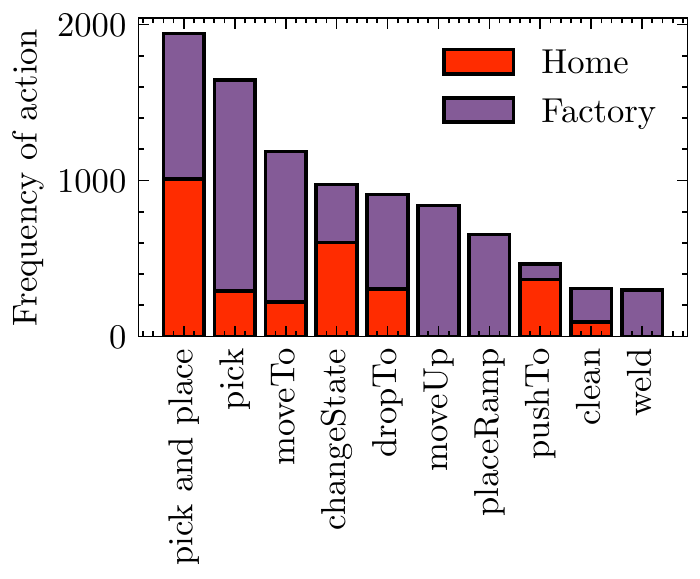}
        \caption{Symbolic actions}
        \label{fig:num_actions}
    \end{subfigure} \setlength{\belowcaptionskip}{-6pt}
    \caption{Data set Characteristics. 
    Distribution of plans with plan length for home and factory domains. Frequency of interaction of top 10 objects and frequency of actions for top 10 actions. 
    The collected data set contains diverse interactions in complex spaces.
    }
    \label{fig:dataset}
\end{figure}

\vspace{0.5ex}
\noindent
\textbf{Annotated Corpus.} 
To curate an imitation learning dataset, we recruit human instructors and provide them with goals. They instruct the robot by specifying a sequence of symbolic actions (one at a time) to achieve each goal. Each action is simulated so that they can observe its effects and the new world state. We encourage the instructors to to complete the task as quickly as possible, making use of available tools in the environment. To familiarize them with the simulation platform, 
we conduct tutorial sessions before data collection. 
%
%
%
%
Our resulting dataset consists of diverse plans with different action sets and object interactions.
We collected plan traces from $\mathrm{12}$ human subjects using  
domain randomization with $\mathrm{10}$ scenes and $8$ semantic goals resulting in 
a corpus of $\mathrm{708}$ and $\mathrm{784}$ plans for the 
home and factory domains. 
%
%
%
%
Figure~\ref{fig:num_int} and \ref{fig:num_actions} show number of interactions with 10 most interacted objects and frequency of 10 most frequent actions respectively. 
The complete set of objects and goals is given in Table~\ref{tab:dataset_complete}.
We also perform data augmentation  by perturbing the metric states in the 
human plan trace, performing random replacements of scene objects and validating plan feasibility in simulation. 
The process results in $\mathrm{3540}$ and $\mathrm{3920}$ plans, respectively. 
%
%
%
Variation was observed in tool use for an instructor for different 
goals, and within similar goals based on context. 
%
%
%
The annotated corpus was split as $(75\%:25\%)$ forming the Training data set and the Test data set  
to evaluate model accuracy. 
A $10\%$ fraction of the training data was used as the 
Validation set for hyper-parameter search. 
No data augmentation was performed on the Test set. 


\vspace{0.5ex}
\noindent
\textbf {Generalization Test Set.}  
%
%
In order to asses the model's capacity to generalize to unseen worlds, we curate a second 
test set environments populated by instances of novel object types 
placed at randomized locations. 
The following sampling strategies were used: 
(i) \emph{Position}: perturbing and exchanging object positions in a scene.
(ii) \emph{Alternate}: removal of the most frequently used tool in demonstrated plans
evaluating the ability to predict the alternative, next best tool to use.
(iii) \emph{Unseen}: replacing an object with a similar object, which is not present in training.
(iv) \emph{Random}: replacing a tool with a randomly picked object which is \emph{unrelated} to the task.
(v) \emph{Goal}: replacing the goals objects. 
This process resulted in a Generalization Test set with $7460$ (goal, plan) pairs.  
%
\section{Experiments}
%
Our experiments use the following accuracy metrics for model evaluation: 
(i) \emph{Action prediction accuracy}: the fraction of tool interactions predicted by the model that 
matched the human demonstrated action $a_t$ for a given state $s_t$, and 
(ii) \emph{Plan execution accuracy}: the fraction of estimated plans that are successful, i.e., can be executed by the robot in simulation and attain the intended goal (in max. $50$ steps). 
%
The implementation and data sets used in experiments  
are available at \url{https://github.com/reail-iitd/tango}. 

\subsection{Comparison with Baselines } 
We compare to the following three baseline models. 
\begin{itemize}
    \item[--] \emph{ResActGraph} model ~\cite{liao2019synthesizing}, 
    augmented with $\mathrm{FastText}$ embeddings~\cite{mikolov2018advances}. 
    \item[--] \emph{Affordance-only} baseline inspired from ~\cite{hermans2011affordance} that learns a 
    co-association between tasks and tools, implemented by 
    excluding the graph convolutions and attention from \modelname{}. 
    \item[--] \emph{Vanilla Deep Q-Learning (DQN)} approach ~\cite{bae2019multi} that 
    learns purely by interactions with a simulator, receiving positive reward for reaching a goal state.
\end{itemize}

Table~\ref{tab:ablation} (top half) compares the \modelname{} model performance with the baseline models. 
The \modelname{} model shows a $14-23$ point increase in \emph{Action prediction accuracy} 
and a $66-71$ points increase in the 
\emph{Plan execution accuracy} when compared to the \emph{ResActGraph} baseline.  
%
%
%
Note that the \textit{ResActGraph} model learns a scene representation assuming a fixed and known set of object types 
and hence can only generalize to new randomized scenes of known objects.  
%
%
In contrast, the \modelname{} model can not only generalize to randomized scenes with known object types 
(sharing the GGCN backbone with \textit{ResActGraph}) but can to novel scenes new object types  
(relying on dense semantic embeddings) and an a-priori unknown number of instances (enabled by a factored likelihood). 

The \emph{Affordance-only} baseline model is confined to learning the possible association between a 
tool object type and the task specified by the human (largely ignoring the environment context). 
This approach addresses only a part of our problem as it ignores the sequential decision making aspect, where 
tools may need to be used in sequence to attain a goal. 

%
Finally, the vanilla DQN baseline achieves less than $20\%$ policy accuracy (even after a week of training). 
In contrast, the \modelname model shows accurate results after training on imitation data for $12-18$ hours. 
The challenges in scaling can be attributed to the problem size ($\approx$1000 actions), long plans, sparse and delayed rewards 
(no reward until goal attainment).  

\begin{table*}[!t]
    \centering
    \resizebox{0.98\textwidth}{!}{
    \begin{tabular}{|c !{\vrule width1pt}c|c !{\vrule width1pt}c|c !{\vrule width1pt}c|c|c|c|c|c|c|}
    \hline 
    \multirow{2}{*}{\textbf{Model}} & \multicolumn{2}{c !{\vrule width0.8pt}}{\textbf{Action Prediction}} & \multicolumn{2}{c !{\vrule width0.8pt}}{\textbf{Plan Execution}} & \multicolumn{7}{c|}{\textbf{Generalization Plan Execution Accuracy}}\tabularnewline
    \cline{2-12} 
     & Home & Factory & Home & Factory & Home (Avg) & Factory (Avg) & Position & Alternate & Unseen & Random & Goal\tabularnewline
    \hline
    \hline
    Baseline (ResActGraph) & 27.67 & 45.81 & 26.15 & 0.00 & 12.38 & 0.00 & 0.00 & 0.00 & 0.00 & 25.10 & 9.12\tabularnewline
    Affordance Only & 46.22 & 52.71 & 52.12 & 20.39 & 44.10 & 4.82 & 17.84 & 47.33 & 29.31 & 29.57 & 34.85\tabularnewline
    DQN  & - & - & 24.82 & 17.77 & 15.26 & 2.23 & 0.00 & 0.00 & 12.75 & 9.67 & 4.21\tabularnewline
    \textbf{\modelname} & 59.43 & 60.22 & 92.31 & 71.42 & \textbf{91.30} & \textbf{60.49} & \textbf{93.44} & \textbf{77.47} & \textbf{81.60} & 59.68 & \textbf{59.41}\tabularnewline
    \hline 
    \multicolumn{12}{|c|}{\textbf{Model Ablations}}\tabularnewline
    \hline 
    - GGCN (World Representation) & 59.43 & 60.59 & 84.61 & 27.27 & 78.02 & 38.70 & 70.42 & 58.79 & 60.00 & 56.35 & 38.64\tabularnewline
    - Metric (World Representation) & 58.8 & 60.84 & 84.61 & 62.34 & 72.42 & 51.83 & 59.68 & 67.19 & 60.79 & \textbf{84.47} & 21.70\tabularnewline
    - Goal-Conditioned Attn & 53.14 & 60.35 & 53.85 & 11.69 & 37.02 & 8.80 & 35.33 & 15.05 & 32.14 & 41.67 & 6.51\tabularnewline
    - Temporal Action History & 45.91 & 49.94 & 24.61 & 0.00 & 8.55 & 0.00 & 0.00 & 0.00 & 0.00 & 30.56 & 1.15\tabularnewline
    - Factored Likelihood & 61.32 & \textbf{61.34} & \textbf{95.38} & \textbf{85.71} & 34.22 & 43.44 & 90.50 & 14.82 & 30.65 & 64.67 & 53.26\tabularnewline
    - ConceptNet & \textbf{63.52} & 60.35 & 89.23 & 57.14 & 81.86 & 56.97 & 82.33 & 68.61 & 74.57 & 65.73 & 47.92\tabularnewline
    \hline 
    \end{tabular}}
    \setlength{\belowcaptionskip}{-6pt}
    \caption{\footnotesize{A comparison of \emph{Action prediction} and \emph{Plan execution} accuracies for the baseline, the proposed \modelname model, and ablations. Results are presented for test and generalization data sets (under five sampling strategies) derived from the home and factory domains. 
    }}
    \label{tab:ablation}
\end{table*}

Next, we assess the zero-shot transfer setting, i.e., 
whether the model can perform common sense generalization in worlds with new objects unseen in training.
The same table shows that the plans predicted 
by \modelname{} lead to an increase of up to $56$ points in plan execution accuracy on Generalization Test set 
over the best-performing baseline model.
This demonstrates accurate prediction and use of unseen tool objects for a given goal. 
%
%
Specifically, in the home domain, 
if the \emph{stool} is not present in the scene, 
the model is able to use a \emph{stick} instead to fetch far-away objects. 
Similarly, if the robot can predict the use of a box for transporting objects even if it 
has only seen the use of a tray for moving objects during training. 
The \textit{ResActGraph} model is unable to adapt to novel worlds and obtains zero points in several generalization tests. 

The poorer performance of the \emph{Affordance-only} model can again be attributed to the fact that 
planning tool interactions involves sequential decision-making. 
Even if the robot can use affordance similarity 
to replace a \emph{tray} object with a \emph{box}, it still needs to predict 
the opening of the \emph{box} before placing an item in its plan for a successful execution. 
This is corroborated by the drop in performance for the \textit{Unseen} generalization tests
for this model by $52.3$ points. 
Finally, the vanilla DQN model lacks a clear mechanism for transferring to novel settings, hence 
shows poor generalization in our experiments. 
\subsection{Ablation Analysis of Model Components} 
We analyze the importance of each component of the proposed model 
by performing an ablation study. 
Table~\ref{tab:ablation} (lower half) presents the results.
For a fair comparison, the model capacities remain the same during the ablation experiments. 

The model builds on the \emph{GGCN} environment representation 
encoding the inter-object and agent-object relational properties. 
The ablation of the GGCN component results in a 
reduction of 22\% in the generalization accuracy in the factory domain 
(where tools may be placed at multiple levels in the factory). 
The inclusion of this component allows the robot to leverage 
relational properties such as OnTop to predict the use of 
tools such as a ramp to negotiate an elevation or a stick to 
fetch an object immediately beyond the manipulator's reach. 

The \emph{Metric} component encodes the metric properties of objects 
in the scene such as positions, size etc. 
Experiments demonstrate its effectiveness in prediction tool interactions based on relative 
sizes of interacting objects. 
E.g., the model successfully predicts that \emph{fruits} can be transported  
using a \emph{tray} but larger \emph{cartons} require a \emph{box} for the same task.
The ablation of this component leads to a reduction of $10.2$ points in the 
\textit{Alternate} generalization tests 
as the ablated model unable to adapt the tool when there are unseen 
objects with different sizes than those seen during training. 

Next, we assess the impact of removing the \emph{Goal-Conditioned Attention component}. 
This experiment shows a a significant reduction ($\approx 50$ points) in 
the \emph{Plan execution accuracy} on the Generalization Test set, 
particularly in scenes with a large number of objects. 
The attention mechanism allows learning of a restricted context of tool objects 
that may be useful for attaining the provided goal, 
in essence, filtering away goal-irrelevant objects populating the scene. 
Additionally, note that the inclusion of this component allows tool predictions 
to be \emph{goal-aware}.
Consequently, we observe ablating this component leads to a reduction of 
$53$ points in the \textit{Goal} generalization test set where 
the goal objects are perturbed.  

%
The \emph{Action History} component utilizes the agent's past interactions 
for the purpose of predicting the next tool interaction. 
The inclusion of this component allows learning of 
correlated and commonly repeated action sequences. 
For instance, the task of exiting from a room typically involves a plan fragment 
that includes moving to a door, opening it and exiting from the door and 
are commonly observed in a number of longer plans. 
The ablation of this component leads to erroneous predictions 
where a particular action in a common plan fragment is missing or incorrectly predicted. 
E.g., a robot attempting to pick an object inside an enclosure without opening the lid. 
In our experiments, we observe that ablating the model 
leads to a significant decrease in goal reach-ability, causing a $70$ point decrease in the 
\emph{Plan Execution accuracy} and $72$ point drop in the \emph{Generalization accuracy}. 

\begin{figure}[!t]
    \centering
    \begin{minipage}{\columnwidth}
        \centering
        \setlength{\belowcaptionskip}{-5pt}
        \includegraphics[width=\textwidth]{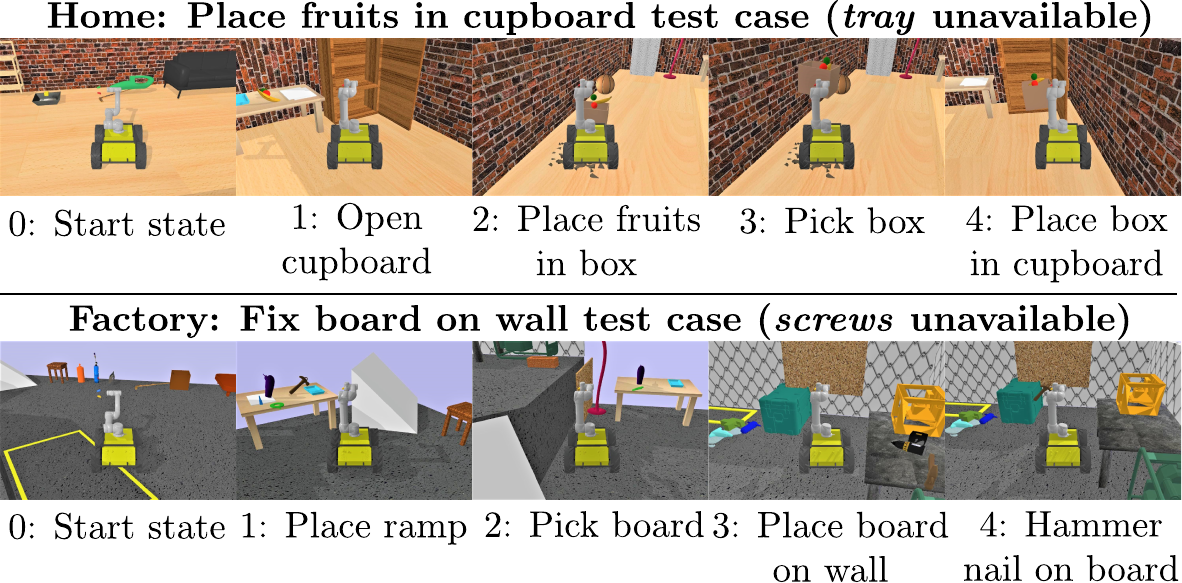}
        \caption{
        \footnotesize{
        A simulated robot manipulator uses \modelname to synthesize tool interactions in 
        novel contexts with unseen objects. 
        }}
        \label{fig:plans}
    \end{minipage}\ 
\end{figure}

\begin{figure}[!t]
        \centering
        \setlength{\belowcaptionskip}{-8pt}
        \includegraphics[width=0.7\linewidth]{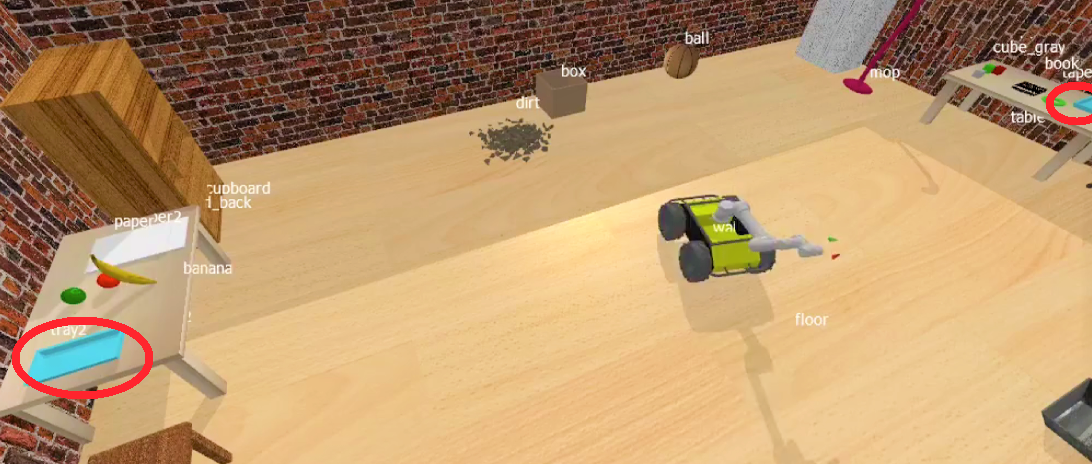}
        \caption{\footnotesize{The model predicts the instance of \textit{tray} (on the left) which is closer to the \textit{fruits} (goal objects) than the one on the right. }} 
        \label{fig:metric_prop}
\end{figure}
\begin{figure}
        \centering
        \setlength{\belowcaptionskip}{-8pt}
        \includegraphics[width=\linewidth]{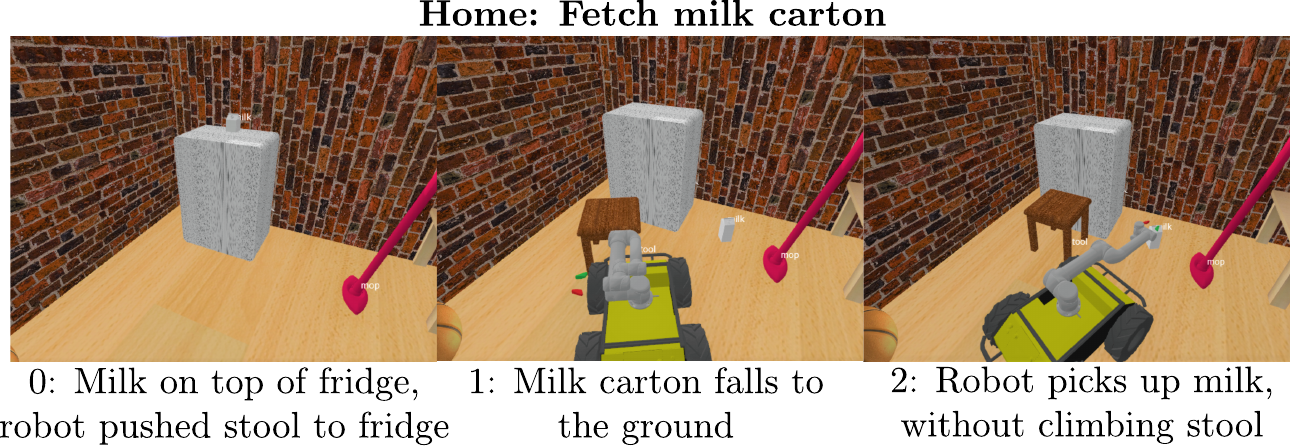}
        \includegraphics[width=0.5\textwidth]{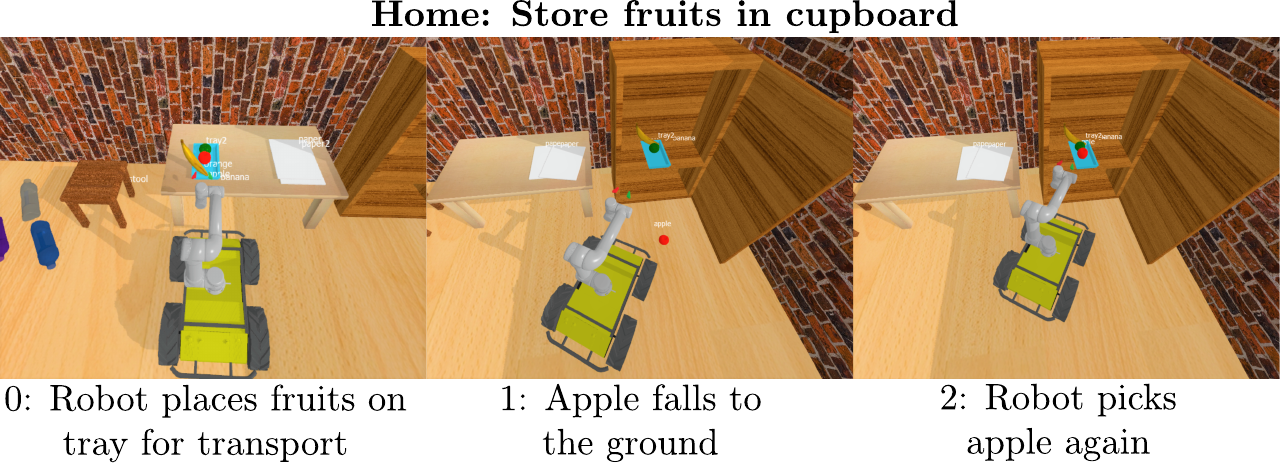}
        \caption{\footnotesize{Interleaved action prediction and 
        execution enables adaptation in case of unexpected errors during action execution. }} 
        \label{fig:robust}
\end{figure}

The need for generalization to novel scenes implies that our model cannot assume  
a fixed and a-priori known set of objects that the robot can interact with. 
Generalization to an arbitrary number of objects in the scene is accomplished by  
by factoring model predictions over individual objects in a recurrent manner. 
Ablating the factored likelihood components results in a simpler model that performs predictions 
over a known fixed-size object set. 
The simplified model displays a higher action-prediction and plan-execution accuracies in 
known world. 
Crucially, we observe that ablating this component results in a significant decrease of   
$51$ and $63$  points in the \textit{Unseen} and the \textit{Alternate} generalization test sets. 
%
 
Finally, the \emph{ConceptNet embeddings} are important 
for semantic generalization to unseen tool types. 
%
%
We replace ConceptNet embeddings with $\mathrm{FastText}$ embeddings to show their importance for the \emph{-ConceptNet} model.
The \emph{-ConceptNet} model (which uses $\mathrm{FastText}$~\cite{mikolov2018advances} word embeddings) shows poorer generalization (6.5\% decrease) as it models
word affinity as expressed in language only. 
$\mathrm{ConceptNet}$ embedding space additionally models relational affinity between objects as maintained in the knowledge-base. 

\subsection{Analysis of Resulting Plans}
Figure \ref{fig:plans} shows the robot using the learned model to 
synthesize a plan for a declarative goal. 
Here, if the goal is to transport fruits and human demonstrates usage of \emph{tray} and the model never sees \emph{box} while training. \modelname uses \emph{box} in a scene where tray is absent, showing that it is able to predict semantically similar tools for task completion. 
Similarly, for the goal of fixing a board on the wall, if humans use \emph{screws} the agent uses \emph{nails} and \emph{hammer} when screws are absent from the scene. 
Figure~\ref{fig:metric_prop} shows how the model 
uses the position information of tool objects to predict the 
tool closer to the goal object or the agent. 
The world representation encodes the metric properties of objects 
(position and orientation) that allows the robot to 
interact with nearer tool objects. 

Figure~\ref{fig:robust} shows the robustness to unexpected errors and stochasticity in action execution.  
Consider the task of ``fetching a carton", where the milk carton is on an elevated platform, 
the model predicts the uses a stool to elevate itself. 
The carton falls due to errors during action execution. 
Following which, the robot infers that the stool is no longer required and directly fetches the carton. 
Similarly, for the task of ``storing away the fruits in the cupboard", the robot predicts the use of tray for the transport task. 
During execution the apple falls off the tray. The robot correctly re-collects the apple.   


\begin{figure}
        \centering 
        \includegraphics[width=0.6\linewidth]{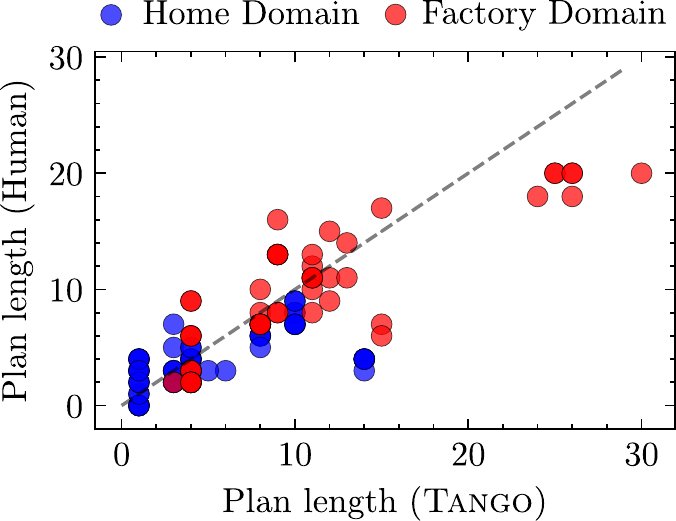}
        \caption{\footnotesize{Scatter plot comparing the lengths of plans obtained from model predictions and those from human demonstrations.}} 
        \label{fig:planlen2}
\end{figure}

Figure~\ref{fig:planlen2} compares the lengths of robot plans 
predicted by the learned model with the set of human demonstrated plans.
We observe that, on average, the predicted plan lengths are 
close to the human demonstrated ones. 
In 12\% cases, the plans predicted by \modelname utilizes 
a more appropriate tool to satisfy the goal condition 
in fewer steps compared to the human demonstrated plan. 
%
%
%
%
%
 
%
\section{Limitations and Future Work}
\begin{figure}[!t]
        \centering 
        \includegraphics[width=0.62\linewidth]{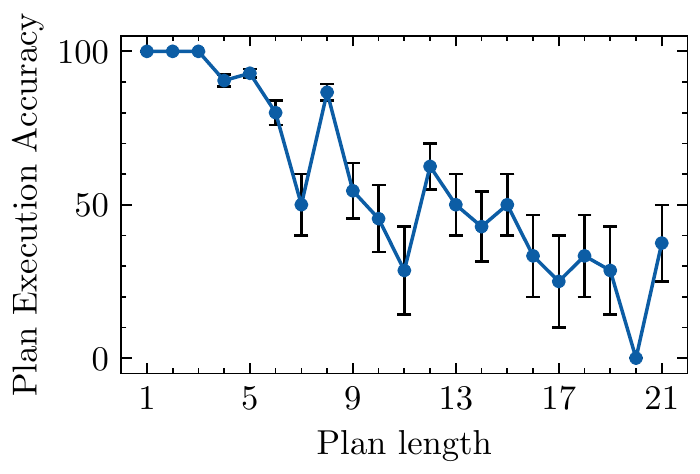}
        \captionof{figure}{\footnotesize{Execution accuracy of inferred plans with plan length.}}
        \label{fig:test}
\end{figure}

\begin{figure}[!t]
        \centering
        \setlength{\belowcaptionskip}{-8pt}
        \includegraphics[width=1.0\linewidth]{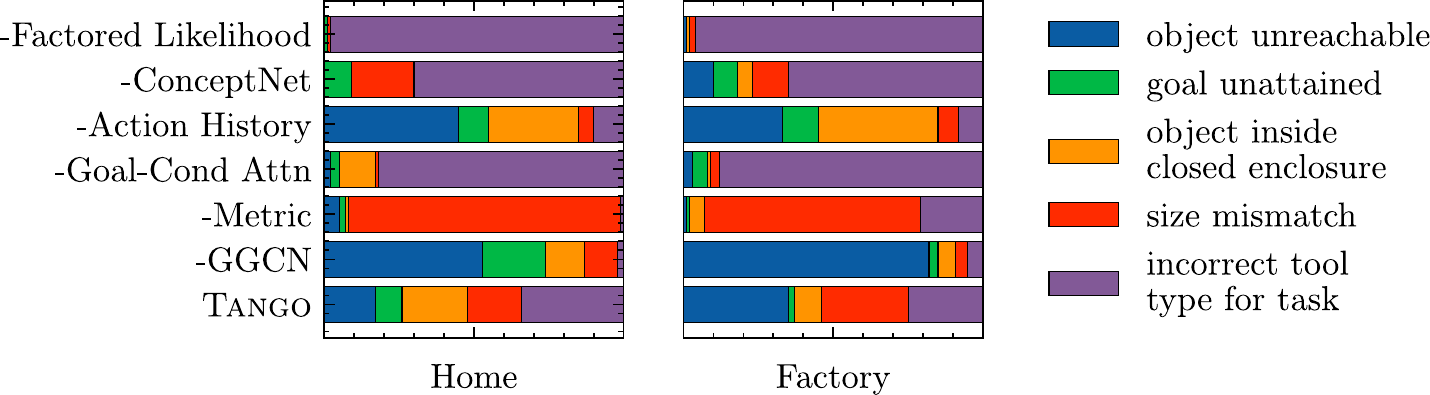}
      \captionof{figure}{\footnotesize{An analysis of fractional errors during plan execution using the learned \modelname model.}}
        \label{fig:errors}
\end{figure}

Figure~\ref{fig:test} assesses the model accuracy 
with the lengths of the inferred plans.  
We observe that the plan execution accuracy decreases by 20\% on the Test sets and 
30\% on Generalization Test sets. 
This merits investigation into planning abstractions \cite{vega2020asymptotically} 
for scaling to longer plan lengths in realistic domains.   
Figure~\ref{fig:errors} analyzes the errors encountered during plan execution using 
actions predicted by the proposed model. 
In 27\% of the cases, the model misses a pre-requisite actions 
required for a pre-condition for initiating the subsequent action. 
For example, missing the need to open the door 
before exiting the room (object unreachable $19\%$) or missing opening the cupboard before picking an object inside it (object inside enclosure $8\%$). 
There is scope for improvement here by incorporating explicit causal structure 
with approaches such as~\cite{nair2019causal}. 
%
%
%
%
Finally, we will explore ways to incorporate the uncertainty 
in the symbolic state (arising from physical sensors) and 
extend the model to partially-known environments; aiding transition 
to experiments on a real platform. 

%% file: conclusions.tex
\section{Conclusions} \label{sec:conclusions}
This paper proposes \modelname, a novel neural architecture that learns a policy to attain intended goals 
as tool interaction sequences leveraging fusion of semantic and metric representations, 
goal-conditioned attention, knowledge-base corpora.
\modelname is trained using a data set of human instructed robot plans with simulated 
world states in home and factory like environments. 
The imitation learner demonstrates accurate commonsense generalization to environments 
with novel object instances using the learned knowledge of shared spatial and 
semantic characteristics. 
It also shows the ability to adapt to erroneous situations and stochasticity in action execution.
Finally, \modelname synthesizes a sequence of tool interactions with a 
high accuracy of goal-attainment.

\section*{Acknowledgements}
{\footnotesize
Mausam is supported by an IBM SUR award, grants by Google, Bloomberg and 1MG, Jai Gupta chair fellowship, and a Visvesvaraya faculty award by Govt. of India. Rohan Paul acknowledges support from Pankaj Gupta Faculty Fellowship and DST's Technology Innovation Hub (TIH) for Cobotics. Shreshth Tuli is supported by the President's Ph.D. scholarship at the Imperial College London. We thank the IIT Delhi HPC facility and Prof. Prem Kalra and Mr. Anil Sharma at the CSE VR Lab for compute resources. 
We express gratitude to the volunteers who participated in the user study. 
We are grateful to anonymous reviewers for their valuable suggestions for improving this paper.
} 

%% file: arxiv2.bbl
\begin{thebibliography}{}

\bibitem[\protect\citeauthoryear{Allen \bgroup \em et al.\egroup
  }{2019}]{allen2019tools}
Kelsey~R Allen, Kevin~A Smith, and Joshua~B Tenenbaum.
\newblock The tools challenge: Rapid trial-and-error learning in physical
  problem solving.
\newblock {\em arXiv preprint arXiv:1907.09620}, 2019.

\bibitem[\protect\citeauthoryear{Bae \bgroup \em et al.\egroup
  }{2019}]{bae2019multi}
Hyansu Bae, Gidong Kim, Jonguk Kim, Dianwei Qian, and Sukgyu Lee.
\newblock Multi-robot path planning method using reinforcement learning.
\newblock {\em Applied Sciences}, 9(15):3057, 2019.

\bibitem[\protect\citeauthoryear{Bahdanau \bgroup \em et al.\egroup
  }{2014}]{bahdanau2014neural}
D~Bahdanau, K~Cho, and Y~Bengio.
\newblock Neural machine translation by jointly learning to align and
  translate.
\newblock {\em arXiv preprint arXiv:1409.0473}, 2014.

\bibitem[\protect\citeauthoryear{Bansal \bgroup \em et al.\egroup
  }{2020}]{toolnet}
Rajas Bansal, Shreshth Tuli, Rohan Paul, and Mausam.
\newblock Toolnet: Using commonsense generalization for predicting tool use for
  robot plan synthesis.
\newblock In {\em Workshop on Advances \& Challenges in Imitation Learning for
  Robotics at Robotics Science and Systems (RSS)}, 2020.

\bibitem[\protect\citeauthoryear{Bisk \bgroup \em et al.\egroup
  }{2020}]{Bisk2020}
Yonatan Bisk, Rowan Zellers, Ronan~Le Bras, Jianfeng Gao, and Yejin Choi.
\newblock Piqa: Reasoning about physical commonsense in natural language.
\newblock In {\em AAAI}, 2020.

\bibitem[\protect\citeauthoryear{Boteanu \bgroup \em et al.\egroup
  }{2015}]{boteanu2015towards}
Adrian Boteanu, David Kent, Anahita Mohseni-Kabir, Charles Rich, and Sonia
  Chernova.
\newblock Towards robot adaptability in new situations.
\newblock In {\em 2015 AAAI Fall Symposium Series}. Citeseer, 2015.

\bibitem[\protect\citeauthoryear{Calli \bgroup \em et al.\egroup
  }{2017}]{calli2017ycb}
B~Calli, A~Singh, J~Bruce, A~Walsman, K~Konolige, S~S Srinavasa, P~Abbeel, and
  A~M Dollar.
\newblock {YCB Benchmarking Project: Object Set, Data Set and Their
  Applications}.
\newblock {\em Journal of The Society of Instrument and Control Engineers},
  56(10):792--797, 2017.

\bibitem[\protect\citeauthoryear{Coumans and Bai}{2016}]{coumans2016pybullet}
Erwin Coumans and Yunfei Bai.
\newblock Pybullet, a python module for physics simulation for games, robotics
  and machine learning.
\newblock {\em GitHub repo}, 2016.

\bibitem[\protect\citeauthoryear{Finn \bgroup \em et al.\egroup
  }{2017}]{finn2017one}
C~Finn, T~Yu, T~Zhang, P~Abbeel, and S~Levine.
\newblock One-shot visual imitation learning via meta-learning.
\newblock {\em arXiv preprint arXiv:1709.04905}, 2017.

\bibitem[\protect\citeauthoryear{He \bgroup \em et al.\egroup }{2015}]{prelu}
Kaiming He, Xiangyu Zhang, Shaoqing Ren, and Jian Sun.
\newblock Delving deep into rectifiers: Surpassing human-level performance on
  imagenet classification.
\newblock In {\em Proceedings of the IEEE international conference on computer
  vision}, pages 1026--1034, 2015.

\bibitem[\protect\citeauthoryear{Hermans \bgroup \em et al.\egroup
  }{2011}]{hermans2011affordance}
Tucker Hermans, James~M Rehg, and Aaron Bobick.
\newblock Affordance prediction via learned object attributes.
\newblock In {\em ICRA: Workshop on Semantic Perception, Mapping, and
  Exploration}, pages 181--184, 2011.

\bibitem[\protect\citeauthoryear{Holladay \bgroup \em et al.\egroup
  }{2019}]{holladay2019force}
Rachel Holladay, Tom{\'a}s Lozano-P{\'e}rez, and Alberto Rodriguez.
\newblock Force-and-motion constrained planning for tool use.
\newblock In {\em IROS}, 2019.

\bibitem[\protect\citeauthoryear{Huang \bgroup \em et al.\egroup
  }{2019}]{huang2019neural}
De-An Huang, Suraj Nair, Danfei Xu, Yuke Zhu, Animesh Garg, Li~Fei-Fei, Silvio
  Savarese, and Juan~Carlos Niebles.
\newblock Neural task graphs: Generalizing to unseen tasks from a single video
  demonstration.
\newblock In {\em CVPR}, pages 8565--8574, 2019.

\bibitem[\protect\citeauthoryear{Jain \bgroup \em et al.\egroup
  }{2015}]{jain2015planit}
Ashesh Jain, Debarghya Das, Jayesh~K Gupta, and Ashutosh Saxena.
\newblock Planit: A crowdsourcing approach for learning to plan paths from
  large scale preference feedback.
\newblock In {\em ICRA}, pages 877--884, 2015.

\bibitem[\protect\citeauthoryear{Kho \bgroup \em et al.\egroup
  }{2014}]{kho2014robo}
Gabriel Kho, Christina Hung, and Hugh Cunningham.
\newblock Robo brain: Massive knowledge base for robots.
\newblock {\em Cornell Univ., USA, Tech. Rep}, 2014.

\bibitem[\protect\citeauthoryear{Kingma and Ba}{2014}]{kingma2014adam}
Diederik~P Kingma and Jimmy Ba.
\newblock Adam: A method for stochastic optimization.
\newblock {\em arXiv preprint arXiv:1412.6980}, 2014.

\bibitem[\protect\citeauthoryear{Kolobov \bgroup \em et al.\egroup
  }{2011}]{kolobov-icaps11}
Andrey Kolobov, Mausam, Daniel~S. Weld, and Hector Geffner.
\newblock Heuristic search for generalized stochastic shortest path mdps.
\newblock In {\em {ICAPS}}, 2011.

\bibitem[\protect\citeauthoryear{Liao \bgroup \em et al.\egroup
  }{2019}]{liao2019synthesizing}
Yuan-Hong Liao, Xavier Puig, Marko Boben, Antonio Torralba, and Sanja Fidler.
\newblock Synthesizing environment-aware activities via activity sketches.
\newblock In {\em CVPR}, pages 6291--6299, 2019.

\bibitem[\protect\citeauthoryear{Mausam and
  Kolobov}{2012}]{kolobov2012planning}
Mausam and Andrey Kolobov.
\newblock Planning with {M}arkov decision processes: An {AI} perspective.
\newblock {\em Synthesis Lectures on Artificial Intelligence and Machine
  Learning}, 6(1):1--210, 2012.

\bibitem[\protect\citeauthoryear{Mikolov \bgroup \em et al.\egroup
  }{2018}]{mikolov2018advances}
T~Mikolov, E~Grave, P~Bojanowski, C~Puhrsch, and A~Joulin.
\newblock Advances in pre-training distributed word representations.
\newblock In {\em LREC}, 2018.

\bibitem[\protect\citeauthoryear{Misra \bgroup \em et al.\egroup
  }{2016}]{misra2016tell}
Dipendra~K Misra, Jaeyong Sung, Kevin Lee, and Ashutosh Saxena.
\newblock Tell me dave: Context-sensitive grounding of natural language to
  manipulation instructions.
\newblock {\em IJRR}, 35(1-3):281--300, 2016.

\bibitem[\protect\citeauthoryear{Nair \bgroup \em et al.\egroup
  }{2019}]{nair2019causal}
Suraj Nair, Yuke Zhu, Silvio Savarese, and Li~Fei-Fei.
\newblock Causal induction from visual observations for goal directed tasks.
\newblock {\em arXiv:1910.01751}, 2019.

\bibitem[\protect\citeauthoryear{Nyga \bgroup \em et al.\egroup
  }{2018}]{nyga2018grounding}
Daniel Nyga, Subhro Roy, Rohan Paul, Daehyung Park, Mihai Pomarlan, Michael
  Beetz, and Nicholas Roy.
\newblock Grounding robot plans from natural language instructions with
  incomplete world knowledge.
\newblock In {\em Conference on Robot Learning}, pages 714--723, 2018.

\bibitem[\protect\citeauthoryear{Puig \bgroup \em et al.\egroup
  }{2018}]{puig2018virtualhome}
Xavier Puig, Kevin Ra, Marko Boben, Jiaman Li, Tingwu Wang, Sanja Fidler, and
  Antonio Torralba.
\newblock Virtualhome: Simulating household activities via programs.
\newblock In {\em CVPR}, 2018.

\bibitem[\protect\citeauthoryear{Speer \bgroup \em et al.\egroup
  }{2017}]{speer2017conceptnet}
R~Speer, J~Chin, and C~Havasi.
\newblock {ConceptNet 5.5: An open multilingual graph of general knowledge}.
\newblock In {\em AAAI}, 2017.

\bibitem[\protect\citeauthoryear{Speer \bgroup \em et al.\egroup
  }{2019}]{conceptnet-github}
R~Speer, J~Chin, and C~Havasi.
\newblock {ConceptNet Numberbatch, the best pre-computed word embeddings you
  can use}.
\newblock {\em GitHub repository}, 2019.

\bibitem[\protect\citeauthoryear{Toussaint \bgroup \em et al.\egroup
  }{2018}]{toussaint2018differentiable}
M~Toussaint, K~Allen, KA~Smith, and JB~Tenenbaum.
\newblock Differentiable physics and stable modes for tool-use and manipulation
  planning.
\newblock In {\em RSS}, 2018.

\bibitem[\protect\citeauthoryear{Vega-Brown and
  Roy}{2020}]{vega2020asymptotically}
William Vega-Brown and Nicholas Roy.
\newblock Asymptotically optimal planning under piecewise-analytic constraints.
\newblock In {\em Algorithmic Foundations of Robotics XII}, pages 528--543.
  Springer, 2020.

\end{thebibliography}
